%% file: paper.tex
\documentclass[letterpaper, 10 pt, conference]{ieeeconf}
\IEEEoverridecommandlockouts    % This command is only needed if
\pdfoutput=1
\input{stachnisslab-latex}

\input{stachnisslab-math}

\newcommand{\name}[0]{{Radar-diffusion}}
\newcommand{\bfvspace}[1]{{\vspace{2mm}\noindent{\textbf{#1}}}}
\usepackage{threeparttable} 
\usepackage{stmaryrd}
\usepackage{overpic}
\title{\LARGE \bf Diffusion-Based Point Cloud Super-Resolution for mmWave Radar Data}

\author{Kai Luan$^*$ \and Chenghao Shi$^*$ \and Neng Wang \and Yuwei Cheng \and Huimin Lu$^\dagger$ \and Xieyuanli Chen$^\dagger$
  \thanks{K. Luan, C. Shi, N. Wang, H. Lu, X. Chen are with College of Intelligence Science and Technology, National University of Defense Technology, China. Y. Cheng is with Tsinghua University and ORCA-TECH.}%
  \thanks{* indicates these authors contribute equally to this work.}
  \thanks{$\dagger$ indicates the corresponding authors: H. Lu (lhmnew@nudt.edu.cn) X. Chen (xieyuanli.chen@nudt.edu.cn)}
  \thanks{This work was partly supported by the National Science Foundation of China under Grant U1913202, U22A2059, and 62203460, Fund for key Laboratory of Space Flight Dynamics Technology (Num 2022-JYAPAF-F1028), Young Elite Scientists Sponsorship Program by CAST (No. 2023QNRC001), and Major Project of Natural Science Foundation of Hunan Province under Grant 2021JC0004.
  }%
}

\begin{document}
\maketitle
\thispagestyle{empty}
\pagestyle{empty}

%%%%%%%%%%%%%%%%%%%%%%%%%%%%%%%%%%%%%%%%%%%%%%%%%%%%%%%%%%%%%%%%%%%%%%%%%%%%%%%%
\begin{abstract}
The millimeter-wave radar sensor maintains stable performance under adverse environmental conditions, making it a promising solution for all-weather perception tasks, such as outdoor mobile robotics.
However, the radar point clouds are relatively sparse and contain massive ghost points, which greatly limits the development of mmWave radar technology.
In this paper, we propose a novel point cloud super-resolution approach for 3D mmWave radar data, named \name{}.
Our approach employs the diffusion model defined by mean-reverting stochastic differential equations~(SDE). Using our proposed new objective function with supervision from corresponding LiDAR point clouds, our approach efficiently handles radar ghost points and enhances the sparse mmWave radar point clouds to dense LiDAR-like point clouds. 
We evaluate our approach on two different datasets, and the experimental results show that our method outperforms the state-of-the-art baseline methods in 3D radar super-resolution tasks.
Furthermore, we demonstrate that our enhanced radar point cloud is capable of downstream radar point-based registration tasks.
\end{abstract}

%%%%%%%%%%%%%%%%%%%%%%%%%%%%%%%%%%%%%%%%%%%%%%%%%%%%%%%%%%%%%%%%%%%%%%%%%%%%%%%%
\section{Introduction}
\label{sec:intro}

Camera and LiDAR are two widely used sensors in robotics and autonomous driving. However, both sensors are vulnerable to adverse weather conditions, such as rain, fog, and snow. With the development of robotics and autonomous driving technologies, there is a great demand for unmanned platforms capable of functioning effectively in harsh environmental scenarios.
Millimeter-wave~(mmWave) radar has received increased attention as it exhibits robust performance in such extreme conditions while providing various measurements of 3D geometric information and additional instantaneous velocity.
% Since this paper focuses on geometrical relevant 3D information, to avoid confusion, we shall exclude the consideration of velocity information throughout the following paper. For clarity, we refer to 3+1D radar as 3D radar and 2+1D radar as 2D radar.
However, radar point clouds suffer from a resolution that is two orders of magnitude lower than LiDAR, presenting significant hurdles for subsequent applications. 
Additionally, radar point clouds are prone to artifacts, ghost points, and false targets due to multipath effects. 
Given the extreme sparsity of radar point clouds, the impact of these noise points is even more pronounced.
Therefore, obtaining denser point cloud data while effectively handling substantial noise points is the pressing research goal for advancing all-weather environmental perception.

Constant false alarm rate (CFAR)~\cite{richards2022fundamentals} is a commonly employed signal processing method for radar, which adjusts the detection threshold based on the background noise, enabling stable detection performance. {However, it struggles to handle a large number of noise points.}
Cheng \etal~\cite{cheng2022tro} propose bypassing CFAR to directly learn extracting high-quality point clouds from raw radar data supervised by LiDAR point cloud.
These methods work on raw radar data, which are exploited to extract high-quality point clouds. However, the extracted point clouds are still sparse.
RadarHD~\cite{prabhakara2023icra} builds dense radar point clouds using an U-Net~\cite{ronneberger2015u}. But it is based on 2D radar point cloud lacking height information and thus cannot handle 3D radar point cloud. 
To the best of our knowledge, no prior super-resolution method for 3D radar point clouds has been proposed.

\begin{figure}[t]
  \centering
  \includegraphics[width=\linewidth]{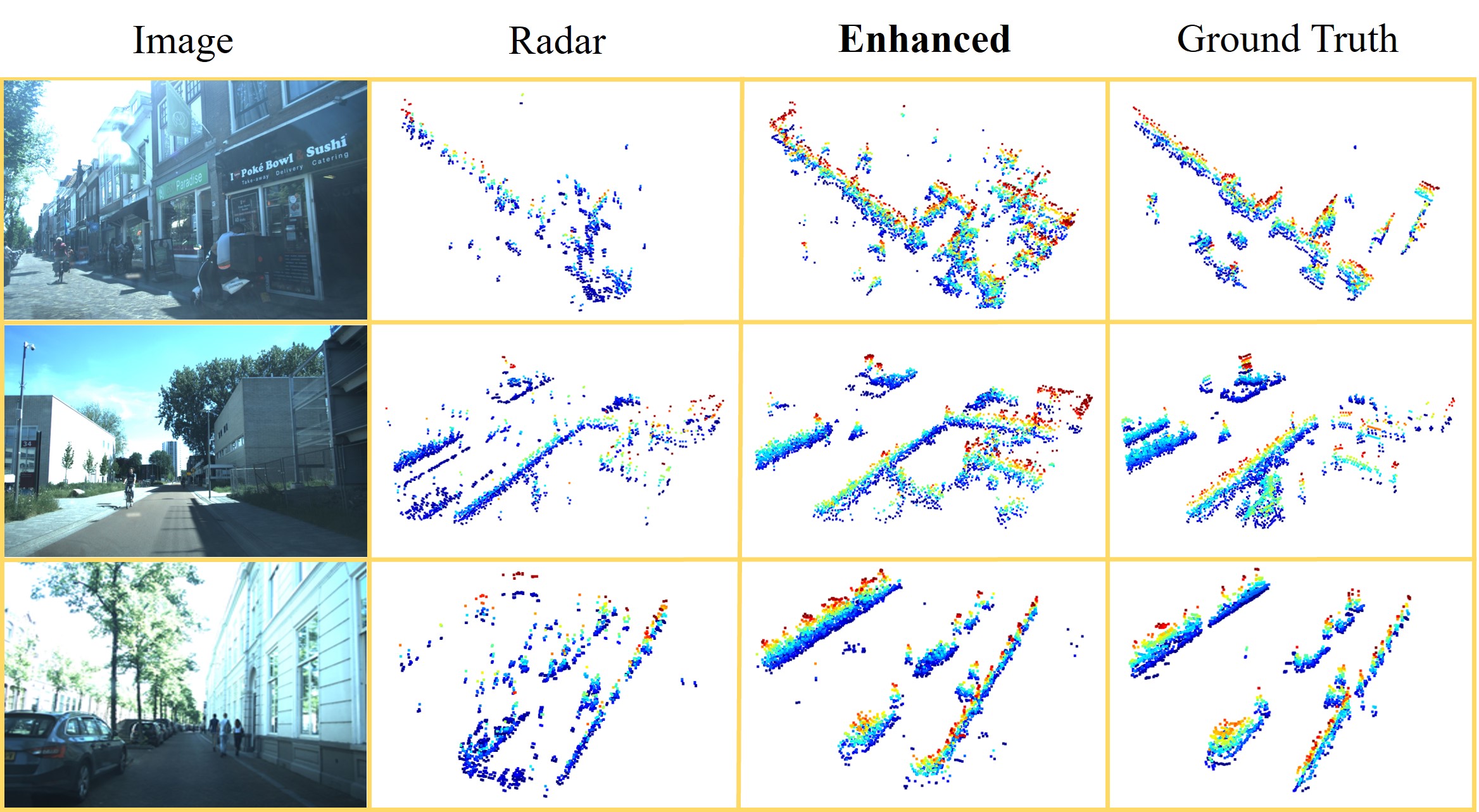}
  \caption{Enhancement effect of our method on radar point clouds. The image, raw radar points, enhanced point clouds, and LiDAR point clouds of the corresponding scene are shown in the figure.}
  \label{fig:motivation}
  % \vspace{-0.4cm}
\end{figure}

Recently, diffusion-based approach, denoising diffusion probabilistic model~(DDPM)~\cite{ho2020nips}, has demonstrated superior performance on image super-resolution~\cite{saharia2022palette} and video restoration~\cite{voleti2022nips}. It generates high-quality super-resolved images by progressively denoising the degraded input image, making the model particularly suited for high-noise scenarios.
However, applying the diffusion model to sparse radar point clouds is still relatively unexplored and challenging.

In this paper, we employ the idea of DDPM and propose a novel \name{} for radar point clouds super-resolution, as shown in \figref{fig:motivation}.
Our approach begins by transforming the radar point clouds into bird's eye view~(BEV) images and then supervised using the corresponding LiDAR BEVs. During training, we use a diffusion model based on mean-reverting stochastic differential equations (SDEs) to process LiDAR BEV images, simulating the transition from denser LiDAR data to radar data using our devised objective function. After training, the model reverse denoising enhances input radar BEV images, producing LiDAR-like, denser results for accurate super-resolution.
We demonstrate that our approach achieves state-of-the-art results in point cloud super-resolution and exhibits robust generalization capabilities to unseen scenarios.
Furthermore, we assess the performance of the generated high-resolution point cloud in the downstream registration task~\cite{shi2021ral,shi2023arxiv}. The results show that the enhanced point cloud can be well used for downstream tasks, revealing its potential for all-weather perception applications.

% {In sum, we make three main contributions}:
% \begin{itemize}
% \item We propose \name{} as the first work to employ the modified diffusion model for achieving dense 3D radar point cloud super-resolution;
% \item The proposed \name{} incorporating with our proposed objective function demonstrates the best performance in 3D radar point cloud super-resolution;
% \item The enhanced point clouds can be well used for downstream registration tasks.
% \end{itemize}

{In summary, our work makes three main contributions}:
\begin{itemize}
\item Proposal of \name{} as the first approach to employ the modified diffusion model for achieving dense 3D radar point cloud super-resolution;
\item Demonstration of the superior performance of the proposed \name{} incorporating our novel objective function in 3D radar point cloud super-resolution;
\item Validation of the usability of the enhanced point clouds for downstream registration tasks.
\end{itemize}

%%%%%%%%%%%%%%%%%%%%%%%%%%%%%%%%%%%%%%%%%%%%%%%%%%%%%%%%%%%%%%%%%%%%%%%%%%%%%%%%
\section{Related Work}
\label{sec:related}

The sparsity and high noise-to-signal ratio of radar point clouds pose critical challenges hindering the development of mmWave radar technology. 
% It is crucial to carefully design relevant algorithms to enhance the resolution of mmWave radar point clouds acquisition and mitigate the interference caused by significant noise. 
Existing approaches for improving the quality of radar point clouds can be categorized into the pre-processing methods and the post-processing methods.

Zhang \etal~\cite{zhang2020mmeye} and Cho \etal~\cite{cho2021guided} propose to replace traditional methods like fast Fourier transform during the signal processing with innovative algorithms~\cite{zhang2020mmeye} or learning-based algorithms~\cite{cho2021guided}.
% One of the main reasons for the sparsity of 3D radar point clouds is the significant information loss caused by CFAR processing.
CFAR~\cite{richards2022fundamentals} is a most commonly employed pre-processing approach. While effectively removing clutter points, CFAR also filters out lots of real detection points, resulting in extremely sparse radar point clouds. 
The learning-based method~\cite{brodeski2019deep} has been proposed as an alternative to the CFAR process, directly operating on range-Doppler images for subsequent tasks. However, these methods require handling a large amount of data, placing high demands on system bandwidth and computational power. Gall \etal~\cite{gall2020spectrum} employ neural networks to estimate the arrival direction of mmWave radar acquisition data, improving the accuracy and enhancing the resolution of acquired point clouds. 
Cheng \etal~\cite{cheng2022tro} propose a radar point detector network for high-quality point cloud extraction incorporating a spatiotemporal filter to handle clutter points. 
However, due to the nature of mmWave radar, target points and clutter points are highly correlated. Introducing more radar detection inevitably causes more clutter points. Furthermore, the generated point clouds using pre-processing methods are still sparse. 

Post-processing methods commonly employ the neural network for clutter point handling and super-resolution. 
Chamseddine \etal~\cite{chamseddine2021icpr} utilize the PointNet~\cite{qi2017cvpr} to distinguish the ghost targets and real targets, resulting in accurate radar point clouds. 
Guan \etal~\cite{guan2020cvpr} effectively recover high-frequency object shapes from the original low-resolution radar point clouds in rainy and foggy weather conditions using a cGAN~\cite{mirza2014arxiv} architecture.
Prabhakara \etal~\cite{prabhakara2023icra} propose the RadarHD employing an U-Net~\cite{ronneberger2015u} to generate LiDAR-like dense point clouds from low-resolution radar point clouds.
Our approach is also a post-processing approach. Unlike existing methods that focus on single-object super-resolution~\cite{guan2020cvpr} or 2D radar point cloud super-resolution~\cite{prabhakara2023icra}, our approach addresses 3D radar point cloud super-resolution within the context of autonomous driving scenes. 
To the best of our knowledge, this is the first approach addressing 3D radar point cloud super-resolution using diffusion model.

\begin{figure*}[t]
  \centering
  \begin{overpic}[width=0.99\linewidth]{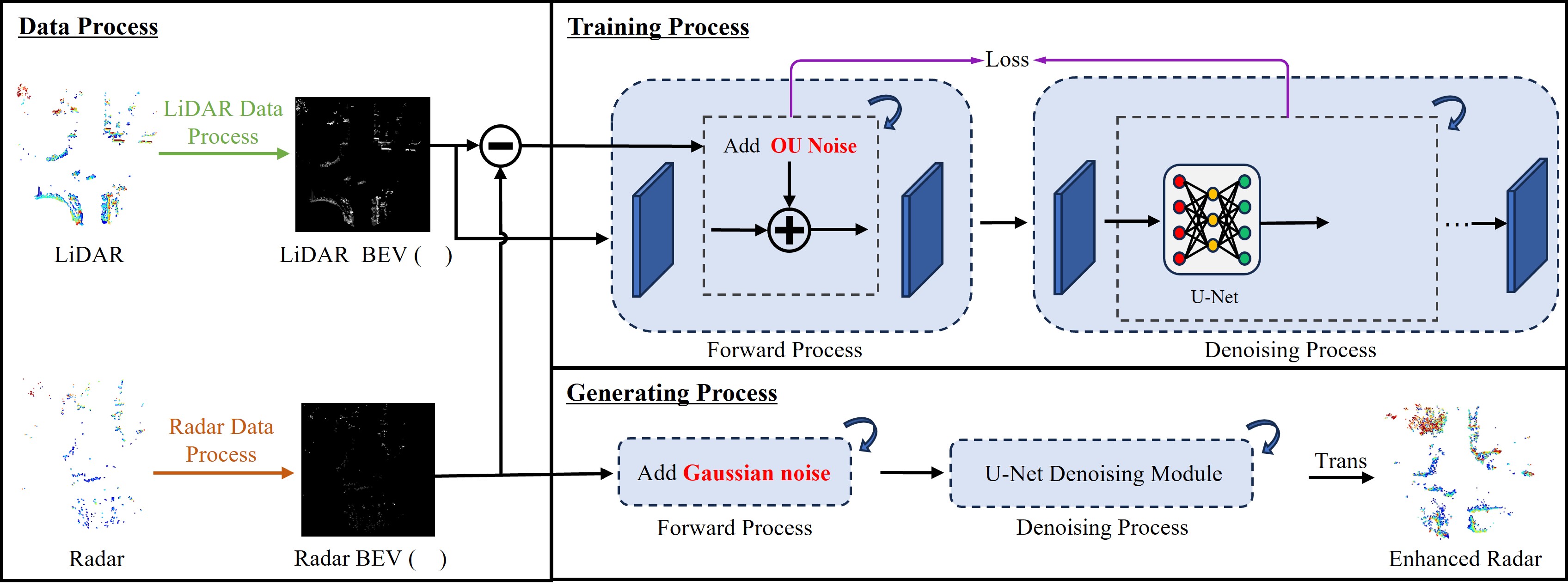}
			% \put(0,20){{\scriptsize ${P}^{ A}\!,\! N^{ A}\!\!\times\!\! 3$}}
   			% \put(8.2,32.2){{$\mu$}}
      %    	\put(17.8,32.4){{$x$}}
      %           \put(28.5,20.5){{$x(0)$}}
      %           \put(46.9,20.5){{$x(T)$}}
      %           \put(60,20.5){{$x(T)$}}
      %           \put(94.5,20.5){{$\tilde{x}(0)$}}
      %           \put(82.6,28.6){{\small$ \tilde{x}(t-1)$}}
      %           \put(47.4,36.8){{$\times T$}}
      %           \put(90.7,36.8){{$\times T$}}
      %           \put(44.7,9.5){{$\times T$}}
      %           \put(76.5,9.5){{$\times T$}}
                % \put(28.8,29.6){{\small{$\mu-x$}}}
                \put(27.2,20.8){{\scriptsize{$x$}}}
                \put(26.8,1.3){{\scriptsize{$\mu$}}}
                \put(40.5,17.2){{\scriptsize{$x(0)$}}}
                \put(57.2,17.2){{\scriptsize{$x(T)$}}}
                \put(67.2,17.2){{\scriptsize{$x(T)$}}}
                \put(95.7,17.2){{\scriptsize{$\tilde{x}(0)$}}}
                \put(85.5,22.6){{\scriptsize{$\tilde{x}(t-1)$}}}
                \put(57.6,30.2){{\scriptsize{$\times T$}}}
                \put(93.8,30.2){{\scriptsize{$\times T$}}}
                \put(55.9,10){{\scriptsize{$\times T$}}}
                \put(81.6,9.8){{\scriptsize{$\times T$}}}
                                
		\end{overpic}
  \caption{Training process and generating process of our proposed \name{}. {The training process models the degradation of LiDAR BEV image to radar BEV image as the forward diffusion process defined by mean-reverting SDE}.
  By learning the reverse denoising process, the LiDAR-like BEV image is then recovered.}
  \vspace{-0.3cm}
  \label{fig:pipeline}
\end{figure*}
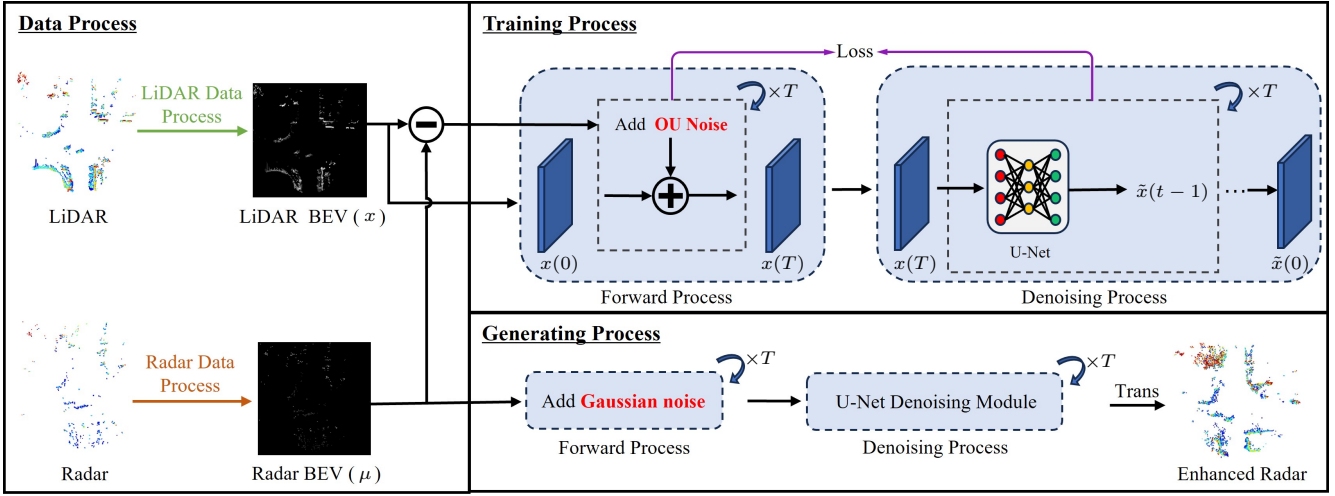

%%%%%%%%%%%%%%%%%%%%%%%%%%%%%%%%%%%%%%%%%%%%%%%%%%%%%%%%%%%%%%%%%%%%%%%%%%%%%%%%
\section{Our approach}
\label{sec:main}

We propose \name{} to enhance sparse mmWave radar point clouds to generate dense LiDAR-like point clouds useful for downstream tasks. The overview of our method is illustrated in \figref{fig:pipeline}.
We begin with converting the radar and LiDAR point clouds into BEV images. 
Subsequently, we model the degradation of high-quality LiDAR BEV images to low-quality radar BEV images using the forward diffusion process of mean-reverting SDE.
By learning the corresponding reverse denoising process using our proposed objective function, high-quality LiDAR-like BEV images are then recovered.
Note that no LiDAR data is required during the generating process. 

%%%%%%%%%%%%%%%%%%%%%%%%%%%%%%%%%%%%%%%%%%%%%%%%%
\subsection{Data Processing}
\begin{figure}[t]
  \centering
  \includegraphics[width=0.99\linewidth]{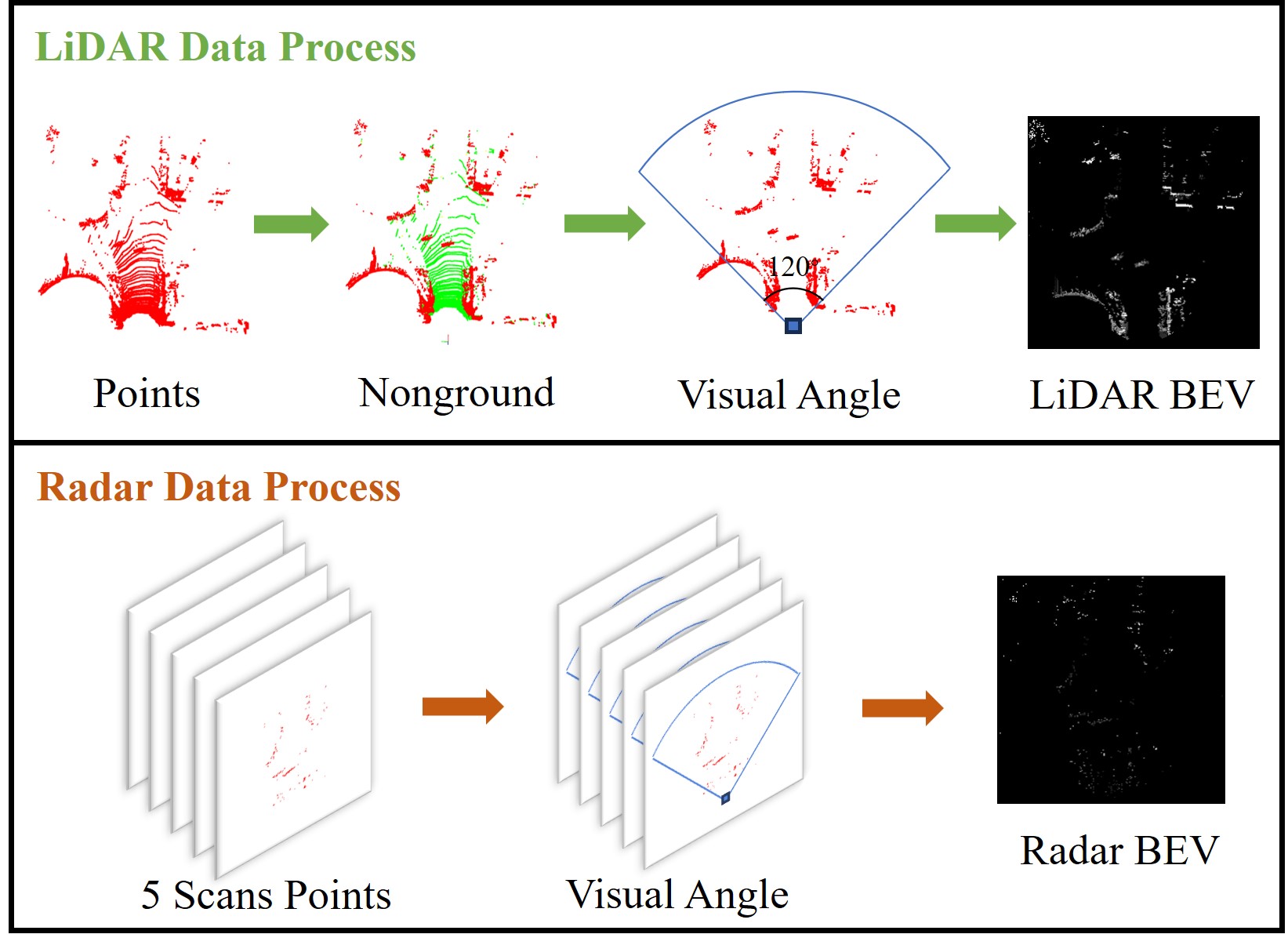}
  \caption{The data process of LiDAR and radar point clouds.}
  \label{fig:dataprocess}
  \vspace{-0.3cm}
\end{figure}

To enable network processing and learning across different sensory modalities, we first convert LiDAR and radar point clouds into BEV images and extract their shared field of view~(FOV).
The overview of the data processing is illustrated in \figref{fig:dataprocess}.

\bfvspace{Ground point removal.} 
We first remove ground points from the raw point cloud data, as they lack valuable semantic information and may hinder the super-resolution learning process.
Furthermore, radar point clouds usually contain few ground points due to the limited resolution of radar echo intensity, so no additional steps are required for their removal. 
For the LiDAR data, we utilize the Patchwork++~\cite{lee2022iros} to detect and remove the ground points from the LiDAR point cloud. It uses adaptive ground likelihood estimation to iteratively approximate the ground segmentation region, ensuring correct separation of ground point clouds even when the ground is elevated by different layers.

\bfvspace{Shared field of view extraction.} 
We then align the LiDAR point $(x_{l}, y_{l}, z_{l})$ to the radar coordinate system by
\begin{equation}
        \begin{bmatrix}
        x_{c} & y_{c}  & z_{c} & 1
        \end{bmatrix}^\top =   \begin{bmatrix}
        \textbf{R}^{r}_{l} & \textbf{t}^{r}_{l}\\
        0  & 1
        \end{bmatrix} \begin{bmatrix}
        x_{l} & y_{l}  & z_{l} & 1
        \end{bmatrix}^\top,
\end{equation}
where $\textbf{R}^{r}_{l}$ and $\textbf{t}^{r}_{l}$ refer to the rotation and translation matrix from the radar to the LiDAR coordinate system.
As we intend to use the BEV image to represent the point cloud, we simplify the shared FOV extraction by focusing solely on their shared horizontal FOV.
We use a Velodyne HDL-64 S3 LiDAR with a horizontal FOV of $360^{\circ}$ and a FRGen21 radar with a horizontal FOV of $120^{\circ}$.
By calculating point yaw angles ($\theta$), we retain LiDAR points and radar points whose yaw angles satisfy $\theta\in[30^{\circ},150^{\circ}]$. 

\bfvspace{BEV generation:} We transform the LiDAR and radar point clouds into compact BEV images with channel information representing height. This facilitates fast and efficient feature learning using mature visual methods. On the other hand, BEV images can better present overall scenes, enabling parallel completion of various perception tasks. To create these BEV images, we retain the points with $x$ coordinates within the range of $[-15,15]$, $y$ within the range of $[0,30]$, and $z$ within the range of $[-0.8,1.7]$. Subsequently, we compress these points into a $256 \times 256$ BEV image with a resolution of $30/256$\,m.
The grayscale value $G_{i,j}$ for each pixel $\{i,j\}$ is determined based on the $z$ value of the highest point falls within that pixel
\begin{equation}
    \begin{aligned}
        G_{i,j} = [ \max (P_{i,j}*\begin{bmatrix}0&0&1\end{bmatrix}^\top) - \gamma ]_+ / {\text{range}_z} * 255,
    \end{aligned}
\end{equation}
where $[\bullet]_+=\max(\bullet,0)$, $P_{i,j}$ represents the point set falls within pixel $\{i,j\}$, and $\gamma$ is a predefined threshold.

\bfvspace{Multi-frame input:} 
Given the sparse nature of the radar point cloud, we combine data from multiple consecutive radar frames using their relative poses. 
In practice, the relative poses can be obtained through the point cloud registration method~\cite{besl1992pami, shi2023tits} or LiDAR odometry method~\cite{chen2019iros}.
We utilize BEV images generated from the aggregated radar point cloud from $5$ consecutive frames as network input.

%%%%%%%%%%%%%%%%%%%%%%%%%%%%%%%%%%%%%%%%%%%%%%%%%
\subsection{{Forward Process based on the Mean-Reverting SDE}}
% To model the degradation of the LiDAR BEV image to the radar BEV image, we utilize the diffusion process defined by mean-reverting SDE.
The standard diffusion process defined by SDE follows
\begin{equation}
    \begin{aligned}
\mathrm{d}x=f({x},t)\mathrm{d}t+g({t})\mathrm{d}w, \quad x(0)\sim p_0(x),
    \end{aligned}
\end{equation}
where $x$ refers to the state linked to LiDAR BEV image, ${f}({x},t)$ and ${g}({t})$ are drift and dispersion functions, and $w$ is a standard Brownian motion.
Typically, this leads to a terminate state $x(T)$ following a Gaussian distribution with zero mean and fixed variance.
Unlike standard SDE widely applied in vision tasks, adding random Gaussian noise to $x$. To model the degradation of the LiDAR BEV image to the radar BEV image, we employ mean-reverting SDE that includes autoregressive Ornstein-Uhlenbeck noise (OU noise)~\cite{lillicrap2015arxiv}  leading to a final state with biased mean and variance. This modification aligns with our objective of matching radar data to LiDAR data, teaching the model how to super-resolve radar data during inference. 
This forward process can be formulated as
\begin{equation}
    \begin{aligned}
\mathrm{d}x=\theta_t\left(\mu-x\right)\mathrm{d}t+\sigma_t\mathrm{d}w, \quad x(0)\sim p_0(x),
    \end{aligned}
\end{equation}
where $\mu$ is the state mean of the radar BEV image, $\theta_t$ characterizes the speed of mean reversion, $\sigma_t$ is the diffusion coefficient.
By setting $\sigma_t^{2}/\theta_t=2\lambda^{2}$, where $\lambda^2$ is the stationary variance, we derive the distribution of $x(t)$ as
\begin{align}
p_t(x)&=\mathcal{N}\big(x(t)\mid m_t(x),v_t\big), 
\label{eq:distribution}
 \end{align}
\begin{align}
m_{t}(x):&=\mu+\left(x(0)-\mu\right)\mathrm{e}^{-\bar{\theta_t}}, 
 \end{align}
\begin{align}
v_t:&=\lambda^2\left(1-\mathrm{e}^{-2\bar{\theta_t}}\right), 
 \end{align}
\begin{align}
\bar{\theta_t}:&=\int_0^t{\theta_z}\mathrm{d}z, 
 \end{align}
where the mean state $m_{t}$ and the variance $v_{t}$ converge to $\mu$ and $\lambda^2$ respectively as $t \to \infty$.
This implies that by progressively adding OU noise, the terminate state of LiDAR BEV image $x(T)$ converges to the radar BEV image $\mu$ with fixed Gaussian noise $\mathcal{N}(0,\lambda)$.

%%%%%%%%%%%%%%%%%%%%%%%%%%%%%%%%%%%%%%%%%%%%%%%%%
\subsection{{Denoising Process on the Mean-Reverting SDE}}
To recover the LiDAR-like BEV image, we reverse the process of mean-reverting SDE by
\begin{equation}
    \begin{aligned}
\mathrm{d}\tilde{x}=\left[\theta_t-\sigma_t^2 \nabla_{\tilde{x}}\log p_t(\tilde{x})\right]\mathrm{d}t+\sigma_t\mathrm{d}w,\,\tilde{x}(T)=x(T),
    \end{aligned}\label{eq:SDE}
\end{equation}
where $\nabla\log p_t(\tilde{x})$ is the score function learned by the time-dependent U-Net~\cite{ho2020nips}.

Specifically, according to \eqref{eq:distribution}, we can obtain the ground truth of $\nabla\log p_t(\tilde{x})$ as
\begin{equation}
    \begin{aligned}
        \nabla_{\tilde{x}}\log p_t(\tilde{x})=-\frac{\tilde{x}(t)-m_t}{v_t}.        
    \end{aligned}\label{eq:gt}
\end{equation}
By rewriting $\tilde{x}(t)=m_t(\tilde{x})+\sqrt{v_t}\epsilon_{t}$, where $\epsilon_{t}\sim\mathcal{N}(0,I)$ is the standard Gaussian noise, we further derive \eqref{eq:gt} as
\begin{equation}
    \begin{aligned}
\nabla_{\tilde{x}}\log p_t(\tilde{x})&=-\frac{\epsilon_t}{\sqrt{v_t}}.
    \end{aligned}
\end{equation}
The neural network predicts the noise $\tilde{\epsilon}(\tilde{x}(t),\mu,t)$ based on current state $\tilde{x}(t)$, condition $\mu$, and time $t$.

%%%%%%%%%%%%%%%%%%%%%%%%%%%%%%%%%%%%%%%%%%%%%%%%%
\subsection{Objective Function}
Instead of using the standard objective that supervises the network to learn the accurate noise, we follow Luo \etal~\cite{luo2023image} and train the U-Net to recover more accurate reversed images by minimizing image residuals at the same stages between forward and denoising processes using the following objective function
\begin{equation}
    \begin{aligned}
J(\tilde{\epsilon}):&=\sum_{i=1}^{T}\gamma_{i}\mathbb{E}\Big[\|\underbrace{\tilde{x}(i)-(\mathrm{d}\tilde{x}(i))_{\tilde{\epsilon}}}_{\mathrm{reversed~}x(i-1)}-x(i-1)\|\Big],\\
&=\sum_{i=1}^{T}\gamma_{i}\mathbb{E}[\|\tilde{x}(i-1)-x(i-1)\|],
    \end{aligned}
\end{equation}
where $\gamma_i$ represents positive weight, $(\mathrm{d}\tilde{x}(i)_{\tilde{\epsilon}}$ represents the mean-reverting SDE defined in \eqref{eq:SDE} using the score $-\tilde{\epsilon}/\sqrt{v_t}$ learned by the network, and $x(i-1)$ is the ideal state for reversed $\tilde{x}(i)$, i.e., the state at time $t=i-1$ in the diffusion process.
This objective function exploits the cumulative error within the denoising process, achieving more stable training for image generation tasks.

However, unlike the visual image generation task, LiDAR and radar BEV images' data distribution is significantly imbalanced.
We observe that the blank area in LiDAR BEV image commonly approximates 20 times larger than the area with actual sensor detection.
Equivalently learning the overall reverse image leads the network to utilize a conservative strategy that simply sets every confused area to blank.
Therefore, we propose dividing the objective function into two parts, considering the blank area and the actual detection area separately.
To this end, we calculate mask matrix $\m{M}=\llbracket x(0)>0\rrbracket$ and $\bar{\m{M}}=\llbracket x(0)==0\rrbracket$, where $\llbracket \bullet\rrbracket$ is an indicator function for which the statement is true.
The modified objective function is written as
\begin{equation}
    \begin{aligned}
J &= J_\text{target} + w \times J_\text{blank}, \\
J_\text{target}& =\sum_{i=1}^{T}\gamma_{i}\mathbb{E}[\|\m{M}\odot\tilde{x}(i-1)-\m{M}\odot x(i-1)\|], \\
J_\text{blank}& =\sum_{i=1}^{T}\gamma_{i}\mathbb{E}[\|\bar{\m{M}}\odot\tilde{x}(i-1)-\bar{\m{M}}\odot x(i-1)\|],
    \end{aligned}\label{eq:object}
\end{equation}
where $\odot$ represents the Hadamard product.
Using our proposed objective function significantly improves the overall performance in our experiments.

\begin{figure*}[t]
  \centering
  \includegraphics[width=0.99\linewidth]{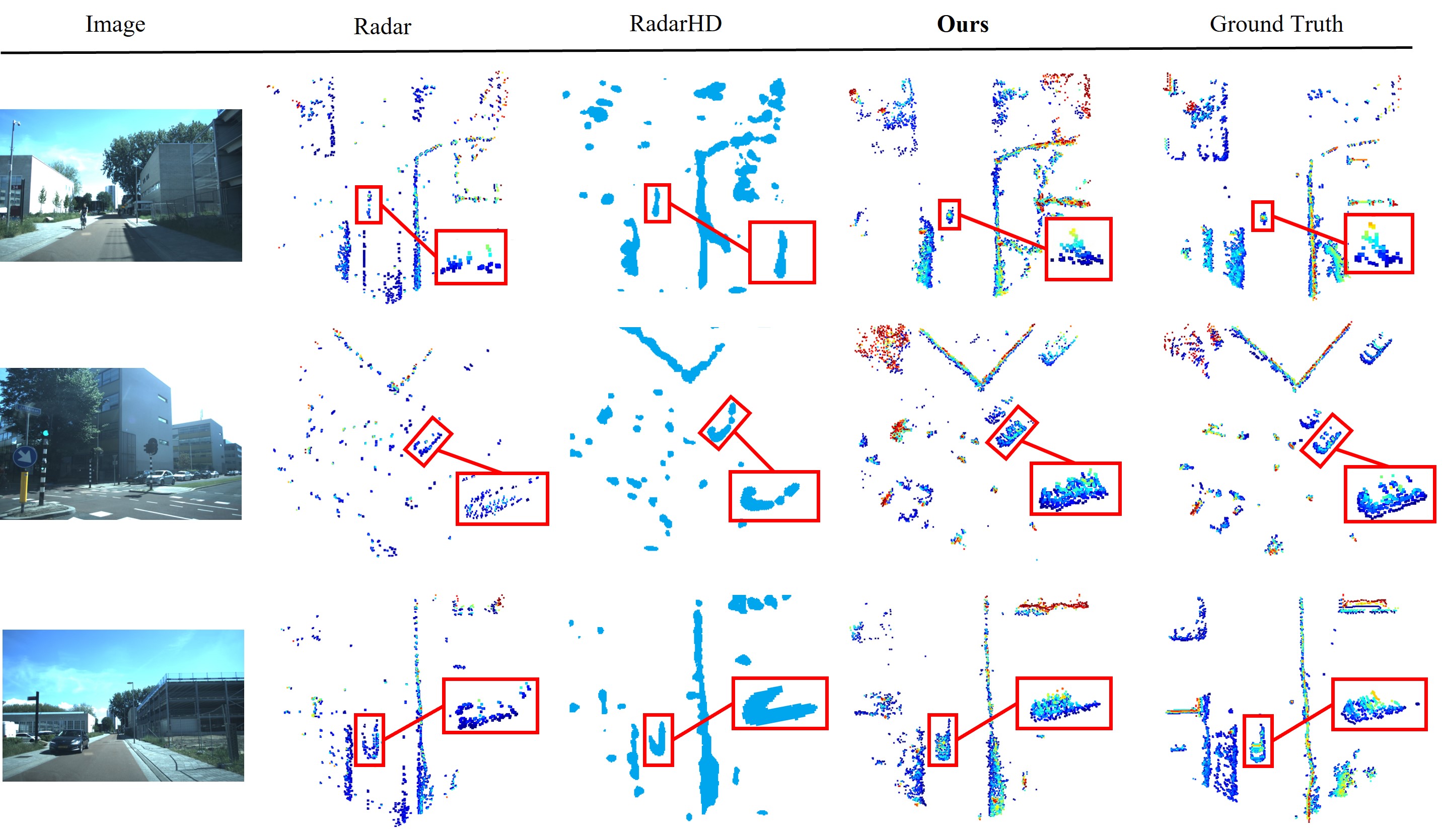}
  \caption{Qualitative results of our method and the RadarHD on the VOD dataset. Specifically, the point clouds enhanced by RadarHD are in 2D, while the point clouds enhanced by our method are in 3D.}
  \label{fig:result}
  \vspace{-0.3cm}
\end{figure*}

%%%%%%%%%%%%%%%%%%%%%%%%%%%%%%%%%%%%%%%%%%%%%%%%%%%%%%%%%%%%%%%%%%%%%%%%%%%%%%%%
\section{Experimental Evaluation}
\label{sec:exp}

%%%%%%%%%%%%%%%%%%%%%%%%%%%%%%%%%%%%%%%%%%%%%%
\subsection{Dataset}
We train and evaluate our approach on View-of-Delft (VOD) dataset~\cite{palffy2022ral} and RadarHD dataset~\cite{prabhakara2023icra}. The VOD dataset contains 8,693 frames of data collected by a Velodyne HDL-64 LiDAR and a FRGen21 radar in complex urban traffic environments.  
The RadarHD dataset is collected in an indoor office environment utilizing Ouster OS0-64 LiDAR and AWR1843 radar without height dimension in the point cloud data. 
Our evaluating datasets cover both outdoor urban roads and indoor environments to test the robustness of different methods.

%%%%%%%%%%%%%%%%%%%%%%%%%%%%%%%%%%%%%%%%%%%%%%%%%
\subsection{Implementation Details}
We train our model using the Lion optimizer~\cite{chen2023arxiv}, with an initial learning rate of $4\times10^{-5}$. 
We choose a more robust noise level $\sigma=50$ and $w=2$ for the forward diffusion process and objective function, respectively. $\gamma_i=1$ is equivalent in all timesteps $t$. We train our model on a single NVIDIA RTX 4090 with a batch size of 8. Total training time takes 9 hours, and the .pth format model size is 306.7\,MB.

%%%%%%%%%%%%%%%%%%%%%%%%
\subsection{Performance on Point Cloud Super-Resolution }
We evaluate the point cloud super-resolution performance of our method on VOD and RadarHD datasets. 
For VOD dataset, we divide it into 7831 frames for training and 635 frames for testing.
The test set contains new frames in unseen sequences, allowing for a comprehensive evaluation of the generalization ability.
% We also conduct experiments on RadarHD dataset for more evaluations.
For RadarHD dataset, we adopt the setup of Prabhakara \etal~\cite{prabhakara2023icra}, utilizing 28 trajectories with 22,784 frames for training and 39 distinct trajectories with 36,779 frames for testing. As the RadarHD dataset only contains 2D radar data, our approach designed for 3D radar point clouds is not directly applicable. Therefore, we employ the following modifications for training on RadarHD dataset. We set the grayscale of the input radar BEV image to the point intensity and the input LiDAR BEV image to $\{0,255\}$, indicating whether there is a point.

\begin{table}[t]
  \caption{Super-resolution Performance.}
  \centering

		\setlength{\tabcolsep}{7.4pt}
\renewcommand\arraystretch{1.2}
  \begin{tabular}{l|c|cc|cc}
    \toprule
			\multicolumn{6}{c}{\textit{VOD dataset}}\\
    \midrule
    % &\multicolumn{1}{c|}{FID} &\multicolumn{3}{c|}{2D} &\multicolumn{3}{c}{One-Way}\\
     & FID$_\text{BEV}\downarrow$ & CD$\downarrow$ 
 & MHD$\downarrow$ & UCD$\downarrow$  & UMHD$\downarrow$ \\ 

    \midrule
    RadarHD  & 247.2 & 0.34  & 0.24 & 0.45 & 0.24 \\
    ours  & \textbf{118.4} & \textbf{0.19} & \textbf{0.10} & \textbf{0.15} & \textbf{0.07} \\

    \toprule
			\multicolumn{6}{c}{\textit{RadarHD dataset}}\\
   \midrule
     & FID$_\text{BEV}\downarrow$ & CD$\downarrow$ 
 & MHD$\downarrow$ & UCD$\downarrow$  & UMHD$\downarrow$ \\ 

    \midrule
    RadarHD  & 141.1 & \textbf{0.44}  & \textbf{0.34} & 0.38  & 0.21 \\
    ours  & \textbf{139.0} & 0.59  & 0.50 & \textbf{0.26} & \textbf{0.13} \\
    
    \bottomrule
  \end{tabular}
  \begin{tablenotes} 
            \item The best results are highlighted in bold. All metrics are in 2D.
    \end{tablenotes} 

  \label{tab:vod}
% \vspace{-0.3cm}
\end{table}

\bfvspace{Metrics}: 
%We employ the following metrics on the enhanced point cloud and LiDAR point cloud for evaluation: 
%The following metrics are employed to evaluate both the enhanced radar point cloud and the LiDAR point cloud:
We employ the following metrics to evaluate the quality of the enhanced radar point cloud comparing to the LiDAR point cloud: (i)~Fréchet Inception Distance~(FID$_\text{BEV}$), the Fréchet distance between the generated enhanced BEV images and LiDAR BEV images. (ii)~Chamfer Distance~(CD), the average distance from each point to the nearest neighbor point in the other point cloud. (iii)~Modified Hausdorff Distance~(MHD), the median distance from each point to the nearest neighbor point in the other point cloud. 
Given that radar possesses a stronger penetration ability, it can detect objects occluded in the LiDAR point clouds, potentially generating more informative point clouds than LiDAR.
We thereby present two more metrics for better evaluation: (iv)~Unidirectional Chamfer Distance~(UCD), the CD from LiDAR point cloud to enhanced radar point cloud only, and (v)~Unidirectional Modified Hausdorff Distance~(UMHD), the MHD from LiDAR point cloud to enhanced radar point cloud only.

\bfvspace{Results}:  
% We test our method against the RadarHD method on the VOD dataset and RadarHD dataset, respectively. Since the RadarHD method can only enhance two-dimensional point clouds, we evaluate the quantitative results of the two methods in generating two-dimensional point clouds in~\ref{tab:vod} and~\ref{tab:RadarHD}. 
Since radar point cloud super-resolution is a new research direction, no similar work currently achieves super-resolution on 3D radar point clouds. The only baseline we find is RadarHD~\cite{prabhakara2023icra} for 2D radar point cloud super-resolution. Thus, for a fair comparison, we conduct all metrics in 2D, \ie, using only $(x, y)$ coordinates, while providing 3D evaluation in our ablation study.
We present experimental results as in~\tabref{tab:vod}.
As depicted on the VOD dataset, our approach produces superior results across all metrics, with an average improvement of 58.4\,\%. Notably, our approach exhibits significant advantages over RadarHD in terms of UCD and UMHD metrics, achieving a 64.7\,\% improvement in UCD and a 70.8\,\% improvement in UMHD. This can be attributed to our approach better faithfully reproduces the information presented in the LiDAR point cloud.
On the RadarHD dataset, our approach maintains its advantages regarding FID, UCD, and UMHD metrics. However, compared to RadarHD, our approach exhibits certain decreases in the CD and MHD metrics. This is because our method can generate denser point clouds and even incorporate information that may not be present in the original LiDAR point cloud. However, the enriched points do not possess real correspondences in the LiDAR point cloud, thus introducing larger errors in CD and MHD metrics. 
The fact that our approach generates denser point clouds is demonstrated in the subsequent qualitative experiments.

To provide more insights into the proposed \name{}, we visualize the enhanced 3D radar point clouds using \name{} as in \figref{fig:result}. As RadarHD is only capable of 2D radar point cloud generation, we visualize its generated BEV image for comparison.
It can be observed that the point clouds enhanced by our method effectively capture the overall layout.
We further zoom in on representative regions, such as vehicles and pedestrians, for closer examination.
As depicted, our enhanced point clouds possess realistic geometric structures for objects while enriching their details that can be occluded in LiDAR point clouds.
% This advantage is reflected in the decrease of metrics CD and MHD in \tabref{tab:vod}, as the enriched points do not possess real correspondences in the LiDAR point cloud, thus introducing larger errors. 

% Due to occlusion issues, LiDAR struggles to capture occluded objects, but it can be observed that the point clouds enhanced using the original radar point clouds contain more occluded objects. Radar has good penetration capabilities, so our enhanced point clouds contain richer scene information than LiDAR point clouds. \figref{fig:result} also shows the magnified effect of the enhanced point clouds, revealing that the enhanced objects have realistic geometric structures and rich detail information. The enhanced point clouds perform well in terms of scene layout accuracy, detail information authenticity, and content richness, making them suitable for downstream tasks such as point clouds registration, object detection, mapping, and localization.

%%%%%%%%%%%%%%%%%%%%%%%%
\subsection{Performance on Downstream Task: Registration }
Our enhanced point clouds present a precise overall layout while possessing enriched details, making them capable of downstream tasks.
In this experiment, we demonstrate the capability of our enhanced point cloud for downstream registration tasks.
We evaluate our method on the test sets of the VOD dataset. 
The point cloud pairs with ground truth pose distances more than $1$\,m are chosen as test samples.

\bfvspace{Metrics}: We employ three metrics to evaluate the registration performance: i) Relative Translation Error~(RTE), which measures the Euclidean distance between estimated and ground truth translation vectors, ii) Relative Rotation Error~(RRE), which is the average difference between estimated and ground truth rotation, and iii) Registration Recall~(RR), representing the fraction of scan pairs with RRE and RTE below certain thresholds, e.g., 5$^\circ$ and 0.5\,m.

\begin{table}[t]
  \caption{Registration Performance using Enhanced Point Cloud.}
  \centering

		\setlength{\tabcolsep}{5pt}
\renewcommand\arraystretch{1.2}
  \begin{tabular}{l|c|cc}
    \toprule
     &RR(\%)$\uparrow$&RTE(m)[succ./all]$\downarrow$&RRE($^\circ$)[succ./all]$\downarrow$  \\ 

    \midrule
    Raw  & 88.51 & 0.11/0.52  & 0.48/2.39  \\
    Enhanced (ours)  & \textbf{93.10} & \textbf{0.11/0.22} & \textbf{0.61/1.13}  \\
    \bottomrule
  \end{tabular}
  \begin{tablenotes} 
            \item The best results are highlighted in bold.
    \end{tablenotes} 

  \label{tab:reg}
\end{table}

\begin{figure}[t]
  \centering
  \includegraphics[width=0.99\linewidth]{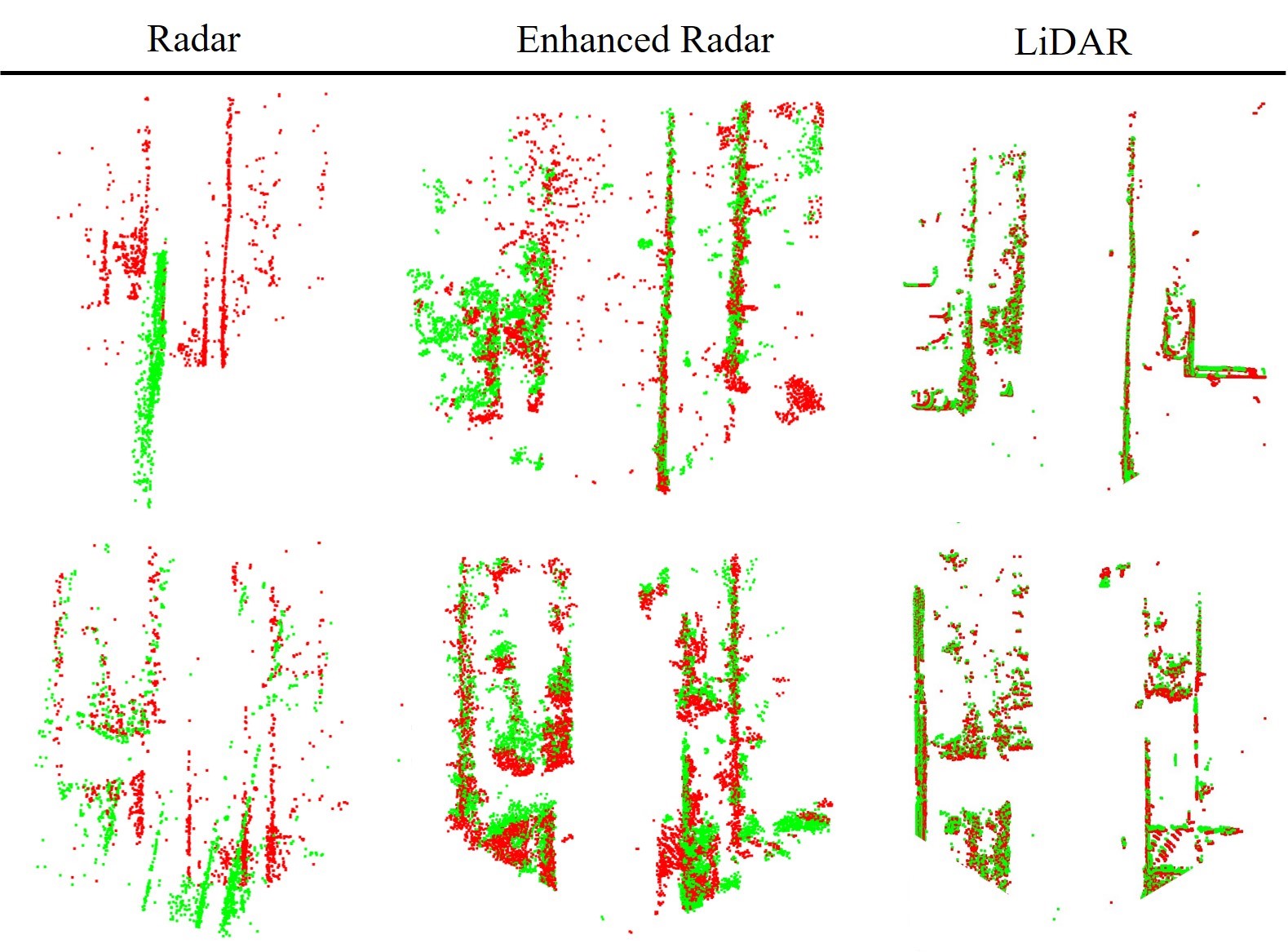}
  \caption{Qualitative results of registration on radar point clouds, our enhanced radar point clouds, and LiDAR point cloud using RDMNet~\cite{shi2023tits}. Different colors represent different frames of point clouds.}
  \label{fig:reg}
\end{figure}

\bfvspace{Results}: In \tabref{tab:reg}, we present registration results on raw radar point clouds and enhanced radar point clouds generated by our \name{} utilizing the state-of-the-art registration method, RDMNet~\cite{shi2023tits}. 
It is a deep learning-based method that finds dense point matches over two point clouds and subsequently performs accurate registration. We directly apply it to different point cloud data to evaluate the registration performance.
% Note that we do not retrain RDMNet but directly employ its official implementation along with an open-source model trained on the LiDAR point cloud.
We adjust the correspondence number according to the point cloud density for a fair comparison. As depicted, our enhanced point clouds exhibit good consistency and accuracy for reliable registration. Furthermore, the more detailed information provided by our enhanced point clouds allows for more robust registration compared to radar point clouds.
We visualize some registration results using different point clouds in \figref{fig:reg}. As can be seen, due to the sparse nature of the original radar data, the registration process failed to align two overlapping radar scans. In contrast, the enhanced radar data can be aligned as well as the registration results of the corresponding LiDAR point clouds.

%%%%%%%%%%%%%%%%%%%%%%%%
\subsection{Ablation Studies} 
We conduct ablation studies to demonstrate the effectiveness of our design in \tabref{tab:ablation}. 
Firstly, we study the objective function.
As shown, compared to the original objective function, using our proposed objective function significantly improves the performance across all metrics. 
Different choices of $w$ in~\eqref{eq:object} have different emphasis on the final performance of the network.
A larger value of $w$ tends to encourage the network to adopt a more conservative approach, making it more inclined to generate unclear or ambiguous regions as blank.
We use $w=2$ as the default as it achieves the most balance performance.
Secondly, we study the number of input frames. As depicted, our approach can work with different numbers of inputs. Merging 5 frames of radar point cloud results in the best performance.
% Since the point clouds from millimeter-wave radar are sparse, we employed multi-frame fusion inputs to add more detailed information, but this also introduced more noisy points. Therefore, we designed ablation experiments to test the enhancement effect of different numbers of frames of point clouds, as shown in \ref{tab:ablation}. Compared to enhancing with a single-frame point cloud input, enhancing with 5-frame radar point clouds resulted in an 11.1\,\%  improvement in Chamfer distance and a 13.4\,\% improvement in Modified Hausdorff distance.

\begin{table}[t]
  \caption{Ablation Study on the VOD-Dataset.}
  \centering

 \setlength{\tabcolsep}{8pt}
\renewcommand\arraystretch{1.2}
  \begin{tabular}{l|c|cc|cc}
    \toprule
    % &\multicolumn{1}{c|}{FID} &\multicolumn{3}{c|}{3D} &\multicolumn{3}{c}{One-Way}\\
    case & FID$_\text{BEV}\downarrow$ & CD$\downarrow$  
 & MHD$\downarrow$  & UCD$\downarrow$   & UMHD$\downarrow$  \\ 

    \midrule
    original  & 165.4 & 1.42  & 1.72 & 2.39 & 1.72 \\
    $w=2$  & 118.4 & \textbf{0.64} & \textbf{0.45} & 0.52 & \textbf{0.32} \\
    $w=4$  & \textbf{103.9} & 0.68 & \textbf{0.45} & 0.73 & 0.42 \\
    $w=5$  & 116.7 & 0.65  & \textbf{0.45} & \textbf{0.51}  & \textbf{0.32} \\

    \midrule
    1scans  & 130.1 & 0.72  & 0.52 & 0.54  & 0.33 \\
    3scans  & 122.1 & 0.67  & 0.47 & \textbf{0.50}  & \textbf{0.31} \\
    5scans  & \textbf{118.4} & \textbf{0.64}  & \textbf{0.45} & 0.52  & 0.32 \\

    \bottomrule
  \end{tabular}
  \begin{tablenotes} 
            \item The best results are highlighted in bold. 
            \item Besides FID$_\text{BEV}$, other metrics are calculated in 3D.
    \end{tablenotes} 

  \label{tab:ablation}
\end{table}

%%%%%%%%%%%%%%%%%%%%%%%%%%%%%%%%%%%%%%%%%%%%%%%%%%%%%%%%%%%%%%%%%%%%%%%%%%%%%%%%
\section{Conclusion}
\label{sec:conclusion}

In this paper, we present a mean-reverting SDE-based diffusion model for 3D mmWave radar point cloud super-resolution.
Our approach models the degradation of high-quality LiDAR BEV image to low-quality radar BEV image as the forward diffusion process, and then learns the reverse process to recover high-quality LiDAR-like BEV images.
We propose to improve the object function to make it more suitable for radar super-resolution tasks. Experiments show that our method can gently handle the massive clutter points in radar point clouds while enhancing them into high-quality LiDAR-like point clouds. Additionally, we demonstrate that our enhanced point clouds can be effectively utilized for downstream registration tasks, laying the foundation for future all-weather perception applications.

%%%%%%%%%%%%%%%%%%%%%%%%%%%%%%%%%%%%%%%%%%%%%%%%%%%%%%%%%%%%%%%%%%%%%%%%%%%%%%%%

\bibliographystyle{plain_abbrv}

\bibliography{glorified,new}

\end{document}

%% file: stachnisslab-latex.tex
\usepackage{graphics}           
\usepackage{times}              
\usepackage{amsmath}            
\usepackage{amssymb}            
\usepackage{graphicx}
\usepackage{algorithm}
\usepackage[noend]{algpseudocode}
\usepackage{booktabs}
\usepackage{color}
\definecolor{instructioncolor}{rgb}{.5,.5,.5}

\usepackage[font=small]{caption}

\def\figref#1{Fig.~\ref{#1}}
\def\tabref#1{Tab.~\ref{#1}}
\def\eqref#1{Eq.~(\ref{#1})}

\makeatletter
\usepackage{xspace}
\DeclareRobustCommand\onedot{\futurelet\@let@token\@onedot}
\def\@onedot{\ifx\@let@token.\else.\null\fi\xspace}
 
\def\ie{i.e\onedot}

\def\etal{{et al}\onedot}
\makeatother

\usepackage{array}
\newcolumntype{L}[1]{>{\raggedright\let\newline\\\arraybackslash\hspace{0pt}}m{#1}}
\newcolumntype{C}[1]{>{\centering\let\newline\\\arraybackslash\hspace{0pt}}m{#1}}
\newcolumntype{R}[1]{>{\raggedleft\let\newline\\\arraybackslash\hspace{0pt}}m{#1}}

%% file: stachnisslab-math.tex
\newcommand{\m}[1]{{\mbox{{\sffamily\slshape{#1\/}}}}}